\documentclass{article}
\usepackage{spconf,amsmath,graphicx,hyperref}
\usepackage{soul,color}
\usepackage{caption}
\usepackage{diagbox}
\usepackage{enumitem}

\makeatletter
\newcommand{\printfnsymbol}[1]{%
  \textsuperscript{\@fnsymbol{#1}}%
}
\makeatother

\title{An end-to-end framework for unsupervised pose estimation of occluded pedestrians}

\name{Sudip Das\textsuperscript{1}, Perla Sai Raj Kishore\textsuperscript{2}, Ujjwal Bhattacharya\textsuperscript{1}}

\address{\textsuperscript{1}Indian Statistical Institute, Kolkata \hspace{0.1cm}
\textsuperscript{2}Institute of Engineering \& Management, Kolkata \hspace{0.1cm}
\\
{\tt\small d.sudip47@gmail.com,  sairajkishore13@gmail.com, ujjwal@isical.ac.in}
}

\begin{document}

%
\maketitle
\begin{abstract}

Pose estimation in the wild is a challenging problem, particularly in situations of (i) occlusions of varying degrees and (ii) crowded outdoor scenes. Most of the existing studies of pose estimation did not report the performance in similar situations. Moreover, pose annotations for occluded parts of human figures have not been provided in any of the relevant standard datasets which in turn creates further difficulties to the required studies for pose estimation of the entire figure of occluded humans. Well known pedestrian detection datasets such as CityPersons contains samples of outdoor scenes but it does not include pose annotations. Here, we propose a novel multi-task framework for end-to-end training towards the entire pose estimation of pedestrians including in situations of any kind of occlusion. 
To tackle this problem for training the network, we make use of a pose estimation dataset, MS-COCO, and employ unsupervised adversarial instance-level domain adaptation for estimating the entire pose of occluded pedestrians. 
The experimental studies show that the proposed framework outperforms the SOTA results for pose  estimation,  instance segmentation and pedestrian detection  in  cases of  heavy  occlusions  (HO)  and reasonable +  heavy occlusions (R + HO) on the two benchmark datasets.
\end{abstract}
%


\section{Introduction}
\label{sec:intro}

Pose estimation of pedestrians is an active area of research with several potential applications.
Research studies in this area have recently achieved significant progress due to the development of different sophisticated deep learning architectures \cite{kocabas2018multiposenet, insafutdinov2016deeper, sun2019deep, xiao2018simple, zhang2019semantics} and several large-scale datasets such as \cite{andriluka20142d, lin2014microsoft, li2019crowdpose}.
However, a majority of the existing related studies did not consider estimating the pose of possible occluded portions of the pedestrians.
It is not an uncommon situation when some major portions of a pedestrian figure remain occluded in the captured view and thus their estimated information about the pose may remain ineffective unless the pose of such occluded parts are explored.  
In fact, studies of pedestrian detection in occluded scenarios has attracted attentions \cite{Zhou2017, Zhang2018, zhang2018occlusion} during the recent years although the scope of these studies are only limited to the detection of pedestrian figures including their occluded portions while excluding estimation of the corresponding pose.
To the best of our knowledge, pose estimation of the occluded portions of pedestrians was considered only in \cite{kishore2019cluenet}.
On the contrary, Ye \textit{et al.} \cite{ye2018occlusion} considered hand pose estimation or Reddy \textit{et al.} \cite{Reddy_2019_CVPR} studied keypoint prediction of 3D shapes of vehicles both for occluded scenarios.



\begin{figure}
               \centering \includegraphics[height=1.75in]{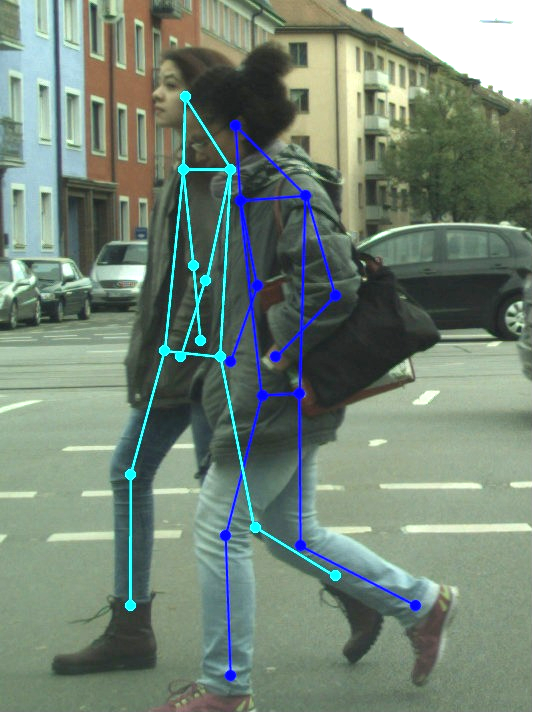}
             \includegraphics[height=1.75in]{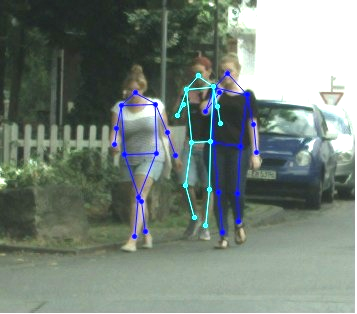}
        \vspace{-0.1in}
        \caption{Pose estimation outputs of the proposed framework for 
        \textcolor{cyan}{occluded} and \textcolor{blue}{completely visible} pedestrian samples appearing in CityPersons dataset.}
        \label{fig:Comparison}
        \vspace{-0.175in}
\end{figure}

\begin{figure*}[!h]
    \centering
    \includegraphics[height=2.7in]{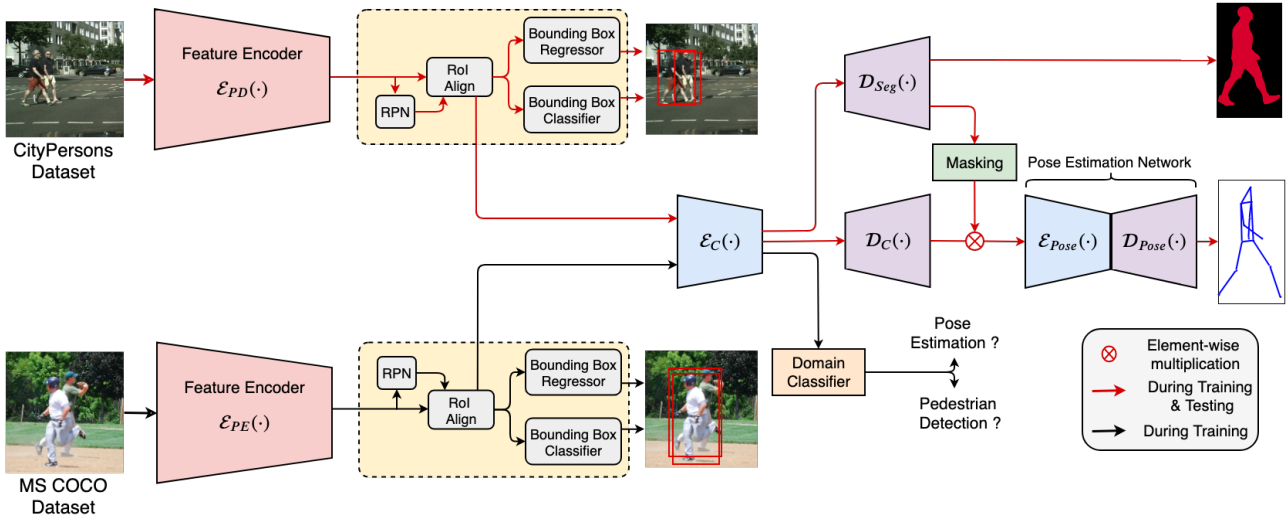}
    \vspace{-0.1in}
    \caption{Architecture of the proposed end-to-end multi-task framework for unsupervised pose estimation of pedestrians.}
    \label{fig:architecture}
    \vspace{-0.19in}
\end{figure*}

In this article, we propose a novel \emph{end-to-end} framework for estimating the entire pose of occluded pedestrians in an unsupervised manner.
This framework tackles the unfavourable situation of the non-availability of two types of training data, such as (i) a pedestrian dataset with pose annotations and (ii) any other dataset with pose annotations for occluded portions of the human figures. 
ClueNet \cite{kishore2019cluenet}, a recently introduced two stage framework, attempted to solve this problem but the same fails to produce desirable results in cases of person-to-person occlusions. On the other hand, the proposed network is equally efficient for every type of occlusion. 


To alleviate the problem of non-availability of required pose annotations for pedestrians from training datasets such as CityPersons \cite{zhang2017citypersons}, we make use of relevant samples of MS-COCO dataset \cite{lin2014microsoft} for pose supervision during the training phase and employ instance-level domain adaptation strategy for this purpose.
The human instances from the pose estimation dataset act as the source domain, whereas the ones from the pedestrian detection dataset act as the target domain for domain adaptation.
The proposed framework is divided into two parts, \textit{domain specific feature extraction} and \textit{domain invariant pose estimation}, which are described in details in the following sections.

In summary, contributions of the present study include 
\begin{enumerate} [noitemsep,nolistsep, wide]
    \item [(1)] A novel \emph{end-to-end framework} for entire pose estimation of occluded pedestrians in an unsupervised manner. To the best of our knowledge, this is the first such an attempt. 
    \item [(2)] Existing pedestrian pose estimation study \cite{kishore2019cluenet} did not consider person-to-person occlusion. However, the present solution is capable of handling all types of occlusions.
    \item [(3)] The proposed strategy has improved SOTA with respect to pose estimation, instance segmentation and pedestrian detection in cases of heavy occlusions (HO) and reasonable+heavy occlusions (R+HO) on MS-COCO and CityPersons dataset.
\end{enumerate}

\section{Proposed Approach}


\subsection{Domain Specific Feature Extraction}

\textit{Domain specific feature extraction} aims to extract robust features of human instances from both pose estimation and pedestrian detection datasets separately, whereas the \textit{domain invariant pose estimation} uses these features to predict the pose of the human instances.
We use an object detection network, Faster-RCNN \cite{ren2015faster}, to extract the features of human instances.
The input image is first passed through a convolutional encoder to generate a dense feature representation.
The Region Proposal Network (RPN) takes these features as input and proposes RoIs where the objects might exist, which are then passed to RoIAlign \cite{he2017mask} for extracting a fixed size feature map for each RoI.
We use RoIAlign instead of originally proposed RoIPool for this purpose and these resized features are finally fed to regression and classification networks for object detection and recognition respectively.

We use two separate detection networks, with feature encoders $\mathcal{E}_{PE}(\cdot)$ and $\mathcal{E}_{PD}(\cdot)$ respectively.
One is used for the pose estimation dataset while the other is used for pedestrian detection dataset. 
Two independent detection networks are necessary as the distribution of the two datasets may not be the same (or even similar), which is usually the case in real world scenarios.
However, using two separate networks allows training in a discrete manner and hence generate strong domain specific features for both datasets.
The intermediate features corresponding to the human instances from the RoIAlign layers of these networks are then passed to the second part for pose estimation.
The architecture of the proposed framework is shown in Figure \ref{fig:architecture}.

\subsection{Domain Invarinat Pose Estimation}

Since the human instances of pedestrian dataset used in this study does not have pose annotations, we use \textit{instance-level domain adaption} to learn to predict their entire pose with the help of human instances in the pose estimation dataset.
The framework, apart from pose supervision from the pose estimation dataset, uses bounding boxes and instance segmentation masks of human figures from both the datasets as an auxiliary information which leads to better pose estimation of the human instances.

Let $\mathcal{X}_{PE}$ be the input images and $\mathcal{Y}_{PE}$ be the corresponding pose annotations of human instances from the pose estimation dataset and similarly let $\mathcal{X}_{PD}$ be the input images from the pedestrian detection dataset.
The RoIAlign features of human instances from both the datasets are first passed through a common convolutional encoder, $\mathcal{E}_{C}(\cdot)$, and are then used for three different purposes -- \textit{instance segmentation}, \textit{domain classification} and finally \textit{pose estimation}.
The purpose of this encoder is to generate \textit{domain invariant features} of the human instances from the extracted domain specific features.
More formally,
\begin{gather}
    \mathcal{F}_{DI}^{(x_{PE}, i)} = \mathcal{E}_{C}(\mathcal{F}_{raw}^{(x_{PE}, i)}),\quad where\; x_{PE} \in \mathcal{X}_{PE}\\
    \mathcal{F}_{DI}^{(x_{PD}, i)} = \mathcal{E}_{C}(\mathcal{F}_{raw}^{(x_{PD}, i)}),\quad where\;x_{PD} \in \mathcal{X}_{PD}
\end{gather}
where $\mathcal{F}_{DI}^{(x_{PE}, i)}$ represents the \textit{domain invariant features} for the $i^{th}$ human instance from image $x_{PE}$ and $\mathcal{F}_{raw}^{(x_{PE}, i)}$ represents the RoIAlign features from the detection network of the $i^{th}$ human instance from image $x_{PE}$. $\mathcal{F}_{DI}^{(x_{PD}, i)}$ and $\mathcal{F}_{raw}^{(x_{PD}, i)}$ are also defined similarly.

For instance segmentation, we pass these \textit{domain invariant features} through a decoder network, $\mathcal{D}_{Seg}(\cdot)$, to generate the segmentation map of the human instances.
We use binary cross-entropy loss to train this part of the network.
Domain adaptation is used to minimize the shift between the two distributions of data, however \textit{instance-level domain adaptation} aims to minimize the shift between particular instances from the data distributions, in contrast to minimizing shift between the data distributions as a whole.
We employ an unsupervised adversarial approach towards this purpose.
The domain invariant features generated by $\mathcal{E}_{C}(\cdot)$ are then passed through a domain classification network, $\mathcal{C}(\cdot)$, to determine to which dataset the human instance features belong to.
The classifier aims to predict the correct dataset, whereas $\mathcal{E}_{C}(\cdot)$ and the network preceding it to confuse the classifier by generating better domain invariant features of the human instances.
This allows the detectors to generate domain specific features for the entire image except for the human instances, whose features are conditioned by the domain classification network.
We use features of human instances from both the datasets, $\mathcal{F}_{DI}^{\mathcal{X}_{PE}}$ and $\mathcal{F}_{DI}^{\mathcal{X}_{PD}}$, for training the domain classification and instance segmentation networks.

For pose estimation, we pass the domain invariant features of the pose estimation dataset, $\mathcal{F}_{DI}^{\mathcal{X}_{PE}}$, through a convolutional decoder, $\mathcal{D}_{C}(\cdot)$, to make the spatial dimensions of the features same as that of the segmentation mask. 
We then multiply (element-wise) these features with the generated segmentation mask and pass them through another encoder-decoder network for pose estimation.
To learn to estimate the entire pose of occluded pedestrians, we make use of the \textit{Mask and Predict} \cite{kishore2019cluenet} strategy which aims to randomly mask a certain percentage of the visible human features used as input and requires the pose estimation network to estimate the pose for both visible as well as the missing parts.
As a result, we mask the segmentation map, generated by $\mathcal{D}_{Seg}(\cdot)$, by certain percentage before multiplying it to the features generated by $\mathcal{D}_{C}(\cdot)$.
Furthermore, since we aim to estimate the entire pose of the human instances, we only use completely visible persons for this purpose. This allows us to have the entire pose of the masked instances as ground truth for back-propagation.
We follow the loss function used in \cite{kocabas2018multiposenet} for training the pose estimation network.
Also, we only use the human instances from the pose estimation dataset for training this part of the network. However, we can use human instances from either of the pose estimation or pedestrian detection dataset during inference.

The overall multi-task loss function, $\mathcal{L}_{total}$, for training the entire framework is given by the weighted combination of losses for the detectors ($\mathcal{L}_{det}$), instance segmentation ($\mathcal{L}_{seg}$), domain classification ($\mathcal{L}_{dc}$) and pose estimation ($\mathcal{L}_{pe}$) as,
\begin{equation}
    \mathcal{L}_{total} = \mathcal{L}_{det} + \alpha\mathcal{L}_{seg} + \beta\mathcal{L}_{dc} + \gamma\mathcal{L}_{pe}
\end{equation}
where $\alpha$, $\beta$ and $\gamma$ are the weights for segmentation, domain classification and pose estimation losses respectively.
Such a multi-task approach allows the networks to learn from each other in a collaborative manner and hence perform better than if they would have been trained separately.



\section{Experiments and Results}

\subsection{Datasets and Training Details}

As mentioned earlier, instance segmentation maps of the input scene images are an indispensable auxiliary information to train the proposed framework for pose estimation.
Due to this constraint, we were able to use only two datasets for this purpose, Citypersons \cite{zhang2017citypersons} for pedestrian detection and MS-COCO \cite{lin2014microsoft} for pose estimation.
We use bounding boxes and instance segmentation masks for the human instances from both the datasets and additionally pose information from MS-COCO dataset.
We used 13 keypoints for pose estimation which include head, left and right elbow, shoulder, wrist, hip, knee and ankle.
And since we are only concerned with human instances we exclusively aim at detecting `person's from both the datasets ignoring all other object categories.

The detector networks were pre-trained on pedestrian detection and pose estimation datasets before training with the second part of the framework. 
We employ Curriculum Learning \cite{bengio2009curriculum} in the \textit{Mask and Predict} strategy by gradually increasing the masking percentage with training.
Momentum Optimizer is used with an initial learning rate of $0.01$ and momentum of $0.9$.
The learning rate is reduced by a factor of $10$ after each subsequent $15k$ iterations.
We set the hyperparameters $\alpha$, $\beta$ and $\gamma$ to 0.5, 1, 1 respectively.
Data augmentation techniques, such as random flipping, blurring, brightness, etc., are employed to further improve the performance of the proposed framework.
All experiments were performed on two Nvidia P6 GPUs.

\begin{figure*}
    \centering
    \includegraphics[height=1.8in]{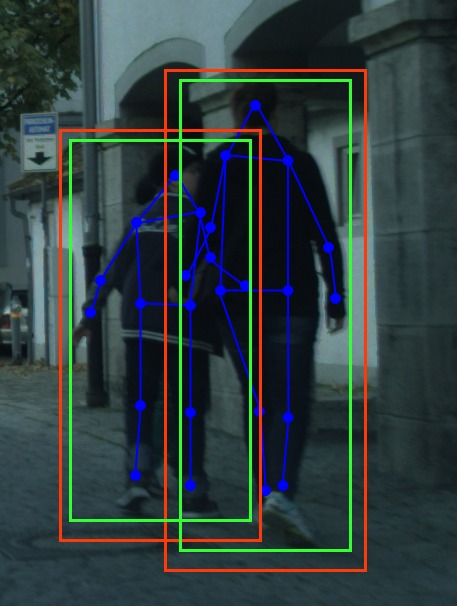}
    \includegraphics[height=1.8in]{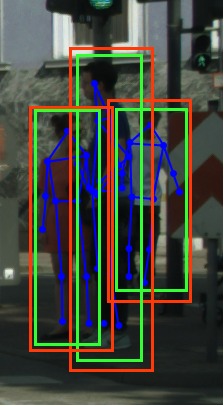}
    \includegraphics[height=1.8in]{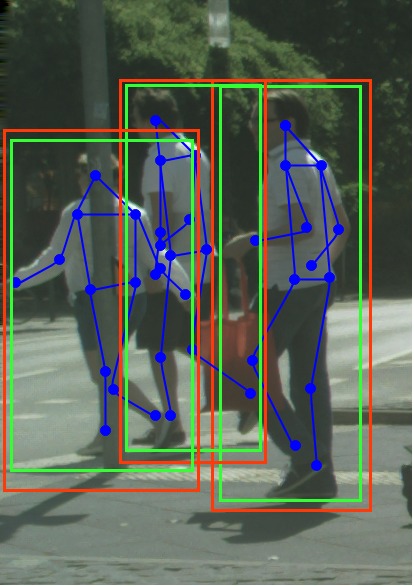}
     \includegraphics[height=1.8in]{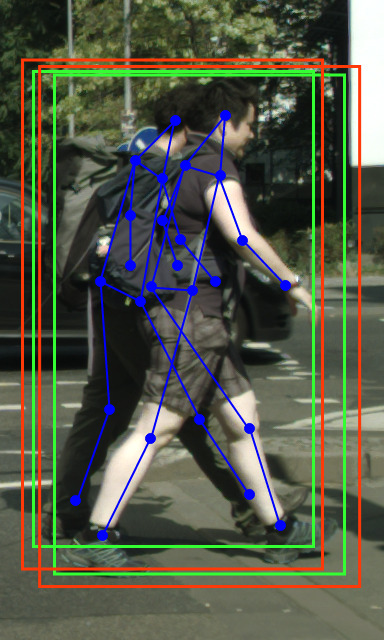}
     \includegraphics[height=1.8in]{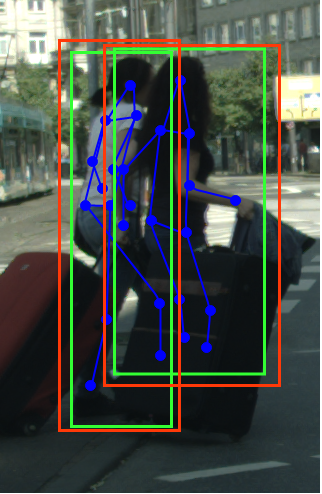} 
     \includegraphics[height=1.8in]{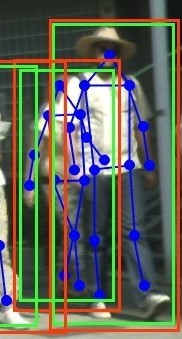}\\
    \vspace{-0.1in}
    \caption{Qualitative results of detection and pose estimation on CityPersons dataset. The \textcolor{red}{ground truth} annotations are shown in red, the \textcolor{green}{detection results} are shown in green and the \textcolor{blue}{predicted pose} is shown in blue.}
    \label{fig:experimental_results}
    \vspace{-.15in}
\end{figure*}

\subsection{Results}



Comparative results of the proposed model on CityPersons and MS-COCO datasets compared with other SOTA methods are provided in this section.
Pose estimation performance of the proposed model on test data of MS-COCO in comparison to a few recently published models have been provided in Table \ref{tab:pose_estimation_results}.
Since we do not have the necessary annotated data to evaluate pose estimation results for cases of occlusions, we artificially created such occluded data by masking the features of certain random parts (by different percentages) of fully visible pedestrians. 
Further, AP scores on similar test data corresponding to different percentages of occlusions provided in Table \ref{tab:pose_estimation_occlusion_results} shows that the proposed method has significantly improved the relevant existing benchmark \cite{kishore2019cluenet}.
Detection results on validation set of CityPersons under different occlusion scenarios have been presented in Table \ref{tab:detection_results}.
Here, we analyzed the detection results for Reasonable + Heavy occlusion \textbf{(R + HO)} with visibility $\in [.20,  \inf]$ and  Heavy occlusion \textbf{(HO)} with visibility $\in [.20, .65]$ and achieved SOTA result in terms of miss rate of \textbf{29.13\%} and \textbf{46.07\%} respectively but for Reasonable occlusion \textbf{(R)} with visibility $\in [.65, \inf]$, our model achieved the third place in terms of miss rate of \textbf{13.29\%} on CityPersons dataset.
Qualitative results of  Detection and Pose Estimation performance have been shown in Figure \ref{fig:experimental_results}.

\begin{table}[!h]
\vspace{-0.1in}
\centering
\caption{AP for state-of-the-art models on MS COCO dataset.}
\vspace{-.10in}
\scalebox{.9}{
\begin{tabular}{|c|c|c|c|c|c|}
\hline
\textbf{Model} & $\textbf{AP}$ & $\mathbf{AP_{50}}$ & $\mathbf{AP_{75}}$ & $\mathbf{AP_{M}}$ & $\mathbf{AP_{L}}$ \\ \hline \hline 

MultiPoseNet \cite{kocabas2018multiposenet}      &  69.6  &   86.3     &  76.6  & 65.0    &  76.3     \\ \hline
CFN  \cite{huang2017coarse}      &  72.6  &   86.7     &  69.7  & 78.3   &  79.0      \\ \hline
 ClueNet\cite{kishore2019cluenet}    &    73.9 &  89.6   &  \textbf{78.2}       &  70.9  & \textbf{79.1} \\ \hline
 \textbf{Ours}    &  \textbf{74.2} &  \textbf{89.9}  & 74.9  &  \textbf{79.3}  & 76.6 \\ \hline
\end{tabular}}
\label{tab:pose_estimation_results}
\vspace{-0.17in}
\end{table}

\begin{table}[!h]
\caption{Results on MS-COCO dataset with different occlusion percentages.}
\centering
\vspace{-.10in}
\scalebox{.8}{
\begin{tabular}{|c|c|c|c|c|c|c|}
\hline
\backslashbox{Model}{Occ. \%} & \textbf{20\%} & \textbf{30\%} & \textbf{40\%} & \textbf{50\%} & \textbf{60\%} & \textbf{70\%} \\ \hline \hline
ClueNet \cite{kishore2019cluenet} & 88.06 & 83.93 & 79.8 & 73.4 & 64.0 & 58.8 \\ \hline
\textbf{Ours} & \textbf{90.3} & \textbf{84.31} & \textbf{81.2} & \textbf{74.06} & \textbf{64.9} & \textbf{59.1} \\ \hline
\end{tabular}}
\label{tab:pose_estimation_occlusion_results}
\vspace{-0.12in}
\end{table}

\begin{table}[!h]
\centering
\caption{Results of instance segmentation on CityPersons.}
\vspace{-0.10in}
\scalebox{.9}{
\begin{tabular}{|c|c|c|c|} 
\cline{1-4}
\textbf{Model}  & \textbf{Training Data}  & \textbf{Person} & \textbf{Rider}   \\
\hline \hline

Mask-RCNN  \cite{he2017mask} & CityPersons + COCO & 34.8 & 27.0\\ 
\cline{1-4}


PANet  \cite{liu2018path} & CityPersons + COCO  & 41.5 & 33.6 \\ 
\cline{1-4}

\textbf{Ours}  & CityPersons + COCO & \textbf{42.1} & \textbf{33.9}\\
\cline{1-4}

\end{tabular}
}
\label{tab:instance_segmentation_results}
\end{table}

\begin{table}[!h]
\centering
\caption{MR for different SOTA models on  CityPersons.}
\vspace{-0.10in}
\scalebox{.8}{
\begin{tabular}{|c|c|c|c|}
\hline
\textbf{Model} & \textbf{R} & \textbf{HO} & \textbf{R+HO} \\ \hline \hline
  Faster RCNN \cite{ren2015faster}   & 15.52  &  64.83  &  41.45    \\ \hline
  Tao et al. \cite{song2018small}  &  14.4  & 52.0   &  34.24   \\ \hline
  OR-CNN \cite{zhang2018occlusion}  &  \textbf{11.0}  & 51.0   &  36.11   \\ \hline
  ClueNet \cite{kishore2019cluenet} & 11.87  & 47.68  &  30.84   \\ \hline
  \textbf{Ours} & 13.29  & \textbf{46.07}  &  \textbf{29.13}   \\ \hline
\end{tabular}}
\label{tab:detection_results}
\vspace{-0.1in}
\end{table}

\begin{table}[!h]

\centering
\caption{Pose estimation results on test data of MS-COCO with different encoder models.}
\vspace{-.10in}
\scalebox{.8}{
\begin{tabular}{|c|c|c|c|c|c|}
\hline
\textbf{Backbone} & $\textbf{AP}$ & $\mathbf{AP_{50}}$ & $\mathbf{AP_{75}}$ & $\mathbf{AP_{M}}$ & $\mathbf{AP_{L}}$ \\ \hline \hline 
VGG-19  \cite{simonyan2014very}    &  63.9  &   80.7     &  70.6  & 58.0   &  70.3      \\ \hline
ResNet-50   \cite{he2016deep}   &  69.1  &   86.1     &  71.6  & 64.2   &  71.1      \\ \hline
ResNet-101  \cite{he2016deep}    &  71.4  &   87.8     &  72.1  & 76.1    &  74.3     \\ \hline
ResNeXt-101 \cite{xie2017aggregated}     &  \textbf{74.2}  &  \textbf{89.9}  &  \textbf{74.9}  & \textbf{79.3}   &  \textbf{76.6}      \\ \hline
\end{tabular}}
\label{tab:different_encoder_results}
\vspace{-0.13in}
\end{table}

Table \ref{tab:instance_segmentation_results} shows the comparative results of instance segmentation on validation set of CityPersons in terms of Average Precision (AP).
The proposed model improves the SOTA with respect to AP score of both person and rider categories.
Pose estimation results (AP scores) of the proposed framework corresponding to different backbone architectures such as VGG-19 \cite{simonyan2014very}, ResNet-50 \cite{he2016deep}, ResNet-101 \cite{he2016deep}, ResNext-101 \cite{xie2017aggregated} have been provided in Table \ref{tab:different_encoder_results}. 
Among these, ResNeXt101 \cite{xie2017aggregated}, is the most efficient encoder network for the present purpose. 


    
      
    

\section{Conclusion}

In this work, we presented a novel end-to-end multi-task framework to precisely estimate the pose of a pedestrian in an unsupervised manner irrespective of its complete visibility or partial occlusions.
We make use of \textit{instance-level domain adaptation} and \textit{instance segmentation masks} to take care of all types of occlusions including person-to-person obstructions.
Experimental results of the proposed strategy on MS-COCO dataset provide strong quantitative evidence of the improvement of pose estimation performance in cases of completely visible and partially occluded human instances over respective state-of-the-art results. 
In future, we shall study towards the development of a pose estimation strategy for occluded humans without the use of segmentation masks.
\bibliographystyle{IEEEbib}
\bibliography{strings,refs}

\end{document}